\documentclass[runningheads]{llncs}
\usepackage[T1]{fontenc}
\usepackage{graphicx,booktabs,tabularx,xcolor,hyperref}
\usepackage{amsmath,subcaption,amsfonts,amssymb,mathtools,booktabs,threeparttable,nccmath}
\usepackage{placeins,url,orcidlink}
\usepackage[title]{appendix}
\hypersetup{colorlinks=true,linkcolor=[rgb]{0.1,0.3,0.9},urlcolor=[rgb]{0.2,0.2,0.2},citecolor=[rgb]{0.1,0.3,0.9}}

\renewcommand{\orcidID}[1]{\nobreak\hspace{0.15em}\orcidlink{#1}}

\makeatletter
\def\input@path{{sections/}{tables/}}
\makeatother

\begin{document}

\title{CATCH: Counterfactual Anatomical Tissue Inpainting with Conditional Haar Diffusion}
\titlerunning{CATCH}

\author{Simon Winther Albertsen \orcidID{0009-0004-0655-6664} \and Hjalte Bjoernstrup \orcidID{0009-0009-7757-4225}
\and Said Djafar Said \orcidID{0009-0009-3570-2057} \and Mostafa Mehdipour Ghazi \orcidID{0000-0002-8473-281X}}

\authorrunning{S. W. Albertsen et al.}

\institute{
Pioneer Centre for AI, University of Copenhagen, Copenhagen, Denmark \\
\email{\{zlp616,fhz806,hdn456\}@alumni.ku.dk, ghazi@di.ku.dk}
}

\maketitle

\begin{abstract}

BraTS local synthesis replaces masked regions in T1-weighted brain MRI with plausible tumor-free tissue while preserving observed anatomy. We present CATCH, conditional 3D diffusion in an invertible Haar-wavelet domain. Its denoiser receives noisy target coefficients, voided-image coefficients, and a signed mask; tumor-excluded wavelet reconstruction and a hole-focused loss guide training, and hard compositing preserves observed voxels. We compare fixed masks, tumor-component augmentation, and a weighted mixture of tumor-derived, irregular-blob, and ellipsoidal masks. Of 25 development cases, five prespecified cases select each arm's checkpoint and all 25 of their trajectory aggregations; a separate 75-case internal set compares the frozen pipelines and selects a weighted mixture for organizer evaluation. Five-trajectory averaging yielded internal SSIM/PSNR/MSE (mean$\pm$SD) of $0.80\pm0.13$, $19.18\pm1.80$\,dB, and $0.010\pm0.005$. As the sole officially evaluated pipeline, weighted mixture yielded $0.772\pm0.119$, $20.89\pm3.27$\,dB, and $0.0098\pm0.0054$ on the 219-case BraTS 2026 validation set. Against compute-matched random augmentation internally, it improved SSIM by 0.019 (95\% bootstrap CI: 0.013--0.025), PSNR by 0.95\,dB, and MSE by 0.003; all three paired comparisons remained significant after Holm correction. Results favor the complete weighted-mixture policy within CATCH; absent official fixed- and random-pipeline scores and a directly comparable external baseline limit broader conclusions.

\keywords{Brain MRI inpainting \and Healthy-tissue synthesis \and Denoising diffusion models \and Wavelet transform \and Medical image synthesis \and BraTS}

\end{abstract}

\section{Introduction} \label{sec:introduction}

Brain tumors distort anatomy and can compromise registration, parcellation, tissue segmentation, and atlas-based pipelines for non-pathological brains \cite{mehdipour2025fast}. Healthy-tissue inpainting replaces pathological regions with plausible tumor-free anatomy for downstream processing. BraTS local synthesis \cite{kofler2023inpainting} requires completing a masked region in a voided T1-native (T1n) volume while preserving observed voxels.

Denoising diffusion probabilistic models (DDPMs) \cite{ho2020ddpm} enable conditional generation \cite{lugmayr2022repaint,saharia2022palette}, but full-resolution 3D diffusion is costly. Wavelet diffusion models (WDMs) \cite{friedrich2024wdm} reduce spatial cost through a fixed invertible transform, and conditional WDM \cite{friedrich2024cwdm} extends them to paired volumetric synthesis. CATCH adapts this framework to study the sampling of artificial healthy-tissue holes.

Supplied masks offer limited spatial and morphological variation. Artificial holes may broaden coverage, but their source, size, shape, placement, and refresh strategy affect the learned reconstruction distribution. We compare three policies within one conditional 3D wavelet-diffusion framework. Our contributions are: (1) a controlled comparison of fixed masks, tumor-component augmentation, and a weighted mixture of tumor-derived, irregular-blob, and ellipsoidal masks, including a compute-matched random-versus-weighted analysis and development-selected trajectory aggregation; (2) full-volume conditional 3D Haar-wavelet diffusion with tumor-excluded supervision and hard preservation of observed voxels; and (3) staged selection: checkpoints on five development cases, inference policies on all 25, and frozen training policies on a separate 75-case set, with paired uncertainty, Holm-corrected tests, and patient-disjoint sensitivity analysis, followed by blind organizer evaluation of the selected pipeline.

\section{Related Work} \label{sec:related-work}

DDPMs learn reverse noising \cite{ho2020ddpm}, while score-based models connect diffusion to stochastic differential equations \cite{song2021scoresde}. RePaint restores known pixels during unconditional sampling \cite{lugmayr2022repaint}; Palette directly conditions its denoiser on the observed image \cite{saharia2022palette}, as does CATCH before final hard compositing. For volumetric synthesis, latent diffusion uses a learned latent space \cite{rombach2022ldm}, whereas WDM applies an exactly invertible 3D wavelet transform \cite{friedrich2024wdm}, and cWDM extends it to paired volumetric synthesis \cite{friedrich2024cwdm}. Diffusion has also addressed 3D healthy-brain inpainting \cite{durrer2024inpainting} in BraTS local synthesis \cite{kofler2023inpainting}, within the broader BraTS benchmark \cite{menze2015brats,bakas2017advancing,baid2021rsna}; fastWDM3D shortens reverse sampling \cite{durrer2026fastwdm3d}. CATCH retains full ancestral sampling and instead studies mask representation, tumor-excluded supervision, online healthy-hole sampling, and trajectory aggregation. Differences in data and evaluation preclude direct numerical comparison with prior scores.

\section{Method} \label{sec:method}

CATCH adapts Tooth-Diffusion \cite{said2025toothdiffusion} from dental CBCT synthesis to brain-MRI inpainting. Let $x_v$ denote the voided volume and $m\in\{0,1\}^{H\times W\times D}$ the complete inpainting region. During training, $m$ is decomposed into disjoint healthy and pathological components, $m=m_h\lor m_u$ and $m_h\odot m_u=0$. The complete image $x$ provides a valid target in $m_h$ but remains pathological in $m_u$, which is excluded from supervision. At inference, only $x_v$ and $m$ are available to the model.

\begin{figure}[!tb]
  \centering
  \includegraphics[width=1\textwidth]{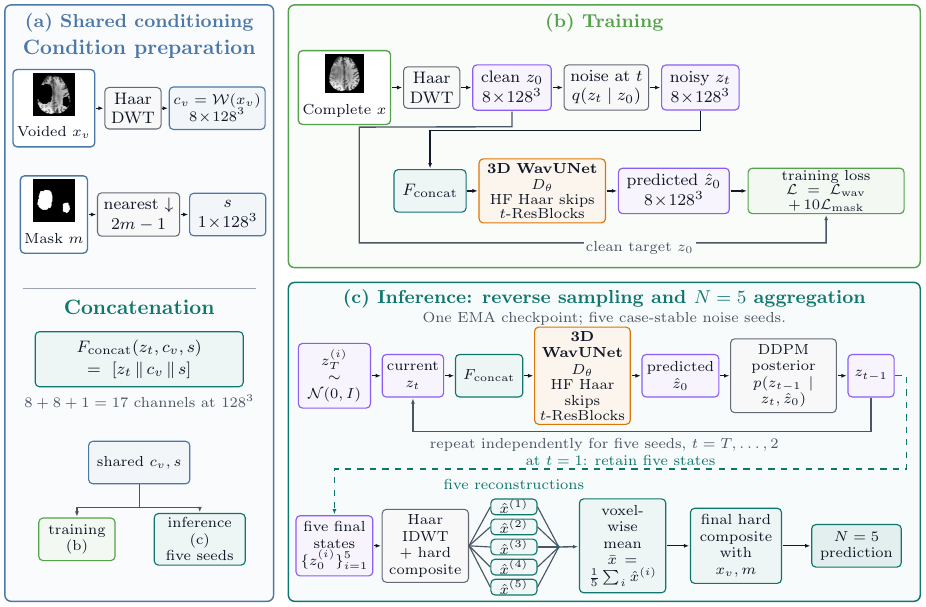}
\caption{CATCH training and inference. A one-level 3D Haar transform produces eight subbands. Noisy target coefficients, voided-image coefficients, and the signed mask form the 17-channel WavUNet input. Predictions are inverse-transformed and hard-composited with the observed image. Random and weighted-mixture pipelines average five case-stable noise trajectories from one exponential-moving-average (EMA) checkpoint; fixed mask uses $N=1$.}

  \label{fig:catch-architecture}
\end{figure}

\subsection{Conditional Wavelet Diffusion} \label{sec:method:wavelet}

Volumes are reoriented to RAS and center-padded from $240\times240\times155$ to $256^3$ without resampling. A one-level 3D Haar transform $\mathcal W$ produces eight $128^3$ subbands. Following WDM \cite{friedrich2024wdm}, the LLL channel is divided by 3 after the forward transform and restored before inversion for target and conditioning coefficients. The transform is therefore fixed and invertible rather than a learned autoencoder.

Let $z_0=\mathcal W(x)$ denote the clean target coefficients. With $T=1000$, $\alpha_t=1-\beta_t$, and $\bar\alpha_t=\prod_{j=1}^{t}\alpha_j$, the forward process is 
\[
q(z_t\mid z_0)=\mathcal N\!\left(z_t;\sqrt{\bar\alpha_t}\,z_0,(1-\bar\alpha_t)\mathbf I\right),
\] where $\{\beta_t\}_{t=1}^{T}$ discretizes the variance-preserving SDE schedule \cite{song2021scoresde} using the continuous rate parameters $\beta_{\min}=0.1$ and $\beta_{\max}=20$. The network predicts $\hat z_0$ directly; ancestral sampling uses the DDPM posterior parameterized by $(z_t,\hat z_0)$ with the \texttt{ModelVarType.FIXED\_LARGE} reverse-variance.

The voided image is transformed as $c_v=\mathcal W(x_v)$. The mask is nearest-neighbor downsampled and encoded as $s=2\,\operatorname{NN}(m)-1\in\{-1,1\}^{128^3}$. The denoiser therefore receives the 17-channel input $\hat z_0=D_\theta([z_t\|c_v\|s],t)$, where $D_\theta$ is a 3D wavelet U-Net \cite{ronneberger2015unet,winther2025rare,friedrich2024wdm}, referred to as WavUNet, with base width 64, channel multipliers (1,2,2,4,4,4), two residual blocks per level, GroupNorm, SiLU, timestep-conditioned scale-and-shift modulation, and wavelet skip connections.

\subsection{Healthy-Tissue Objective} \label{sec:method:objective}

Target and conditioning image share the inference-available scale $a=\max(x_v)$. Intensities are divided by $a$, clipped to $[0,1]$, and mapped to $[-1,1]$. Let $e_{k,p}=(\hat z_0^{(k)}(p)-z_0^{(k)}(p))^2$ denote the error in subband $k$ at wavelet location $p\in\Omega$. We set $u_p=0$ if the corresponding $2^3$ voxel cell intersects $m_u$, and $u_p=1$ otherwise:
\begin{equation}
\mathcal L_{\mathrm{wav}}=
\frac{1}{8|\Omega|}
\sum_{k=1}^{8}\sum_{p\in\Omega}u_p e_{k,p}.
\label{eq:wavelet-loss}
\end{equation}
This matches the implemented fixed-grid normalization: cells touching the real tumor contribute zero error but remain part of the fixed denominator. Because the denominator does not shrink with the excluded region, $\mathcal L_{\mathrm{wav}}$ implicitly down-weights the per-voxel training signal for cases with larger tumors.

For the hole-focused term, cells intersecting $m_h$ form a binary coverage map that is Gaussian-smoothed with $\sigma=2$ and multiplied by $u$ to obtain $w_p\ge0$:
\begin{equation}
\mathcal L_{\mathrm{mask}}=
\frac{\sum_{k=1}^{8}\sum_{p\in\Omega}w_p e_{k,p}}
{8\sum_{p\in\Omega}w_p},
\quad
\mathcal L=\mathcal L_{\mathrm{wav}}+10\mathcal L_{\mathrm{mask}}.
\label{eq:training-objective}
\end{equation}
When $\sum_p w_p>0$, the implementation uses Eq.~\eqref{eq:training-objective}; if this support is zero, it assigns that example's regional term zero rather than adding an epsilon. Both terms exclude cells touching the real tumor and use coefficient-space mean-squared error, consistent with the direct clean-coefficient regression objective in the adapted wavelet-diffusion implementation. We fix this objective across arms to isolate the healthy-mask policy.

\subsection{Healthy-Mask Training Variants} \label{sec:method:augmentation}

The \emph{fixed-mask} variant uses the supplied $m_h$. Online variants sample $m_h'$, void $m_h'\cup m_u$, and supervise $m_h'$ while excluding $m_u$. Their tumor-component bank was built from connected components of the corresponding raw BraTS-GLI 2023 tumor-segmentation labels, rather than the supplied enlarged $m_u$ masks, containing only the 1,151 optimization cases, with five repeat slots per case.

\paragraph{\textbf{Random augmentation.}}
This reference-style policy uses tumor components only. Current-case components are allowed; source masks are sampled by inverse tumor-to-brain-ratio rank within a conditioned 10\% window, transformed by the reference flips and rotations, and fixed across epoch runs. Exhausted placement raises an error rather than silently changing the policy.

\paragraph{\textbf{Weighted-mixture augmentation.}}
This policy samples 80\% tumor components, 15\% irregular blobs, and 5\% ellipsoids. Current-case tumor components are excluded; empirical target volumes, bounded 3D rotations, optional anisotropic scaling, and epoch-wise slot refresh are used. Accepted masks are connected, contain 800--225,070 voxels, remain at least five voxels from $m_u$, and overlap background by at most 25\%. Exhausted generation falls back to a nonempty supplied healthy mask that does not overlap $m_u$.

\begin{figure}[!t]
  \centering
  \includegraphics[width=1\textwidth]{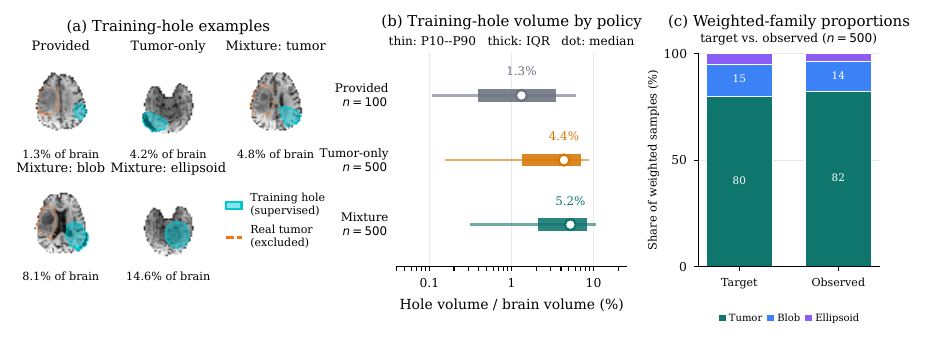}
  \caption{Training-mask audit. (a) Supplied and generated masks; cyan and orange denote supervised voxels and excluded tumor. (b) Epoch-0 hole-volume distributions (10th--90th percentile, interquartile range, and median; $n=100/500/500$). (c) Target and observed weighted-mixture proportions: tumor/blob/ellipsoid $80/15/5\%$ and $82.2/14.0/3.8\%$ (411/70/19); no fallback occurred.}
  \label{fig:mask-augmentation-audit}
\end{figure}

\subsection{Sampling and Compositing} \label{sec:method:inference}

We use exponential-moving-average (EMA) weights and the full 1000-step ancestral reverse process. At each step, \texttt{clip\_denoised=True} inverse-transforms the predicted clean state, clips it in normalized voxel space, and returns it to the wavelet domain. The final reconstruction is restored to native scale, shape, and orientation and hard-composited as $\hat x=x_v\odot(1-m)+\hat x_{\mathrm{gen}}\odot m$.

Global seed 0 and the case- and member-stable \texttt{brats-validation-noise-v1} scheme make trajectories shard-independent and preserve the first members across larger $N$. Multi-trajectory policies independently decode and composite each sample, aggregate volumes voxel-wise, and composite once more. Thus, trajectories come from one checkpoint rather than independently trained models.

\section{Experiments} \label{sec:experiments}

\subsection{Data and Split Protocol} \label{sec:exp:data}

The ASNR-MICCAI BraTS Local Synthesis release \cite{kofler2023inpainting,menze2015brats,bakas2017advancing,baid2021rsna,karargyris2023medperf,akbari2017segmentation,bakas2017tcgalgg} contains 1,251 training cases with complete and voided T1n volumes, a complete inpainting mask, and healthy and unhealthy submasks.

A deterministic 100-case hold-out (seed 2026) leaves 1,151 optimization cases, excluded from the hold-out loader and component bank. The hold-out is split into 25 development cases for within-arm checkpoint and inference-policy selection and 75 internal model-selection cases for comparing the frozen training arms. Patients do not overlap between these two subsets, but the split is case-keyed: 3/25 development and 11/75 model-selection cases have sibling timepoints in the optimization pool. None of the five checkpoint-diagnostic cases has such a sibling. We therefore report a post-hoc analysis on the 64 model-selection cases that are patient-disjoint from optimization.

The three models differ only in healthy-mask policy and use seed 0, batch size one, AdamW \cite{loshchilov2019adamw} with learning rate $10^{-5}$ and zero weight decay, linear decay over a 300k-step horizon, EMA decay 0.9999, and FP32 training on NVIDIA A100 40\,GB GPUs.

\subsection{Checkpoint, Inference, and Training-Policy Selection} \label{sec:exp:selection}

\paragraph{\textbf{Checkpoint selection.}}
Checkpoints are saved every 5k steps through 295k, with a terminal EMA save at step 299,999. Every 10k steps, one deterministic 1000-step trajectory is evaluated on five prespecified development cases; fixed-mask step 200k is unavailable. Within each arm, checkpoints are ranked by aggregate SSIM, PSNR, and MSE, retaining the lowest mean rank. These diagnostics serve only checkpoint selection, not final performance estimation.

\paragraph{\textbf{Trajectory-policy selection.}}
At each terminal EMA state, all 25 development cases compare nested $N\in\{1,3,5\}$ trajectories under voxel-wise mean or median aggregation. Each selected policy is transferred to the corresponding checkpoint and frozen before the 75-case comparison, assuming the preferred aggregation is stable across late checkpoints.

\paragraph{\textbf{Training-policy selection.}}
The three frozen arm-specific pipelines are evaluated once on the separate 75-case set. The pipeline with the lowest mean joint rank across SSIM, PSNR, and MSE is selected for organizer validation; no checkpoint or inference setting is retuned on these cases.

\subsection{Evaluation and Statistical Analysis} \label{sec:exp:evaluation}

The official evaluator normalizes prediction and target to $[0,1]$ using the voided-image 0.5th and 99.5th percentiles and scores only the supplied healthy submask. We report SSIM, PSNR, and MSE. Means and sample standard deviations describe case heterogeneity, not training-seed variability.

For case $i$ and pipeline $j$, let $r_{ij}^{\mathrm{SSIM}}$, $r_{ij}^{\mathrm{PSNR}}$, and $r_{ij}^{\mathrm{MSE}}$ denote ranks among the frozen pipelines. Their mean joint rank is 
\[
\textstyle R_j=(3n)^{-1}\sum_{i=1}^{n}(r_{ij}^{\mathrm{SSIM}}+r_{ij}^{\mathrm{PSNR}}+r_{ij}^{\mathrm{MSE}}),
\]
with $n=75$ for internal training-policy selection; development configurations are ranked analogously within each training arm. For each metric, a Friedman test assesses the omnibus difference and two-sided paired Wilcoxon tests assess the three pairwise contrasts, with Holm correction within metric. Mean paired differences use deterministic 100k-resample bootstrap 95\% CIs (seed 2026). Predictions, per-case metrics, split audits, hashes, and run metadata are retained in CSV/JSON format. Code and scripts are available at \url{https://github.com/simonwinther/brats2026}.

\section{Results} \label{sec:results}

\subsection{Development Selection} \label{sec:results:selection}

Fixed-mask steps 170,000 and 180,000 tied for the lowest mean rank; we retained step 180k because it led in PSNR and MSE. Random step 290,000 and weighted-mixture step 200,000 led all aggregate metrics within their respective histories.

Table~\ref{tab:ensemble-selection} summarizes trajectory-policy selection on all 25 development cases. Mean $N=5$ had the lowest joint rank for both augmented models. For fixed mask, mean $N=3$ improved the rank over $N=1$ by only 0.133 at triple trajectory cost, whereas mean $N=5$ was worse. We therefore retained the fixed mask with $N=1$ and both augmented models with mean $N=5$.

\begin{table}[t]
  \centering
  \caption{Development-set joint rank at the terminal EMA states. Bold and underlining mark the lowest and second-lowest values within each arm. Fixed-mask $N=1$ was retained for the compute--performance trade-off.}
  \label{tab:ensemble-selection}
  \small
  \setlength{\tabcolsep}{8pt}
  \begin{tabularx}{\textwidth}{@{}Xccc@{}}
    \toprule
Configuration & Fixed mask $\downarrow$ & Random aug. $\downarrow$ & Weighted mixture $\downarrow$ \\
    \midrule
    $N=1$ & \underline{2.920} & 4.707 & 4.400 \\
    Mean $N=3$ & \textbf{2.787} & \underline{2.133} & 2.693 \\
    Median $N=3$ & 2.973 & 3.533 & 3.773 \\
    Mean $N=5$ & 3.240 & \textbf{1.600} & \textbf{1.587} \\
    Median $N=5$ & 3.080 & 3.027 & \underline{2.547} \\
    \bottomrule
  \end{tabularx}
\end{table}

\subsection{Internal 75-Case Training-Policy Selection} \label{sec:results:confirmation}

Table~\ref{tab:model-selection-results} summarizes the frozen pipelines on the internal model-selection set. Omnibus differences were significant for SSIM ($\chi_F^2(2)=52.83$), PSNR, and MSE (both $\chi_F^2(2)=52.91$; all $p<3.4\times10^{-12}$). Weighted mixture led all three metrics and had the lowest joint rank, so it was selected for organizer evaluation. Relative to fixed mask, it improved SSIM by 0.041 (95\% CI: 0.032--0.050), PSNR by 1.34\,dB (1.08--1.61), and $[0,1]$-scale MSE by 0.0038 (0.0029--0.0047). Because the selected fixed pipeline used $N=1$, this primary comparison was not compute-matched; a post-hoc all-$N=5$ analysis is reported below.

\begin{table}[t]
  \centering
  \caption{Internal 75-case training-policy selection results (mean$\pm$SD). Lower joint rank is better. Bold and underlining mark the best and second-best results per column. The $N=1$ pipeline is not compute-matched to the $N=5$ pipelines.}
  \label{tab:model-selection-results}
  \small
  \setlength{\tabcolsep}{2.0pt}
  \begin{tabularx}{\textwidth}{@{}Xcccc@{}}
    \toprule
    Pipeline & SSIM $\uparrow$ & PSNR (dB) $\uparrow$ & MSE $\downarrow$ ($\times10^{-2}$) & Rank $\downarrow$ \\
    \midrule
    Fixed ($N=1$) & 0.756$\pm$0.162 & 17.84$\pm$1.75 & 1.365$\pm$0.807 & 2.551 \\
    Fixed (mean $N=5$) & 0.756$\pm$0.162 & 17.85$\pm$1.75 & 1.364$\pm$0.807 & 2.551 \\
    Random (mean $N=5$) & \underline{0.778$\pm$0.140} & \underline{18.24$\pm$2.19} & \underline{1.252$\pm$0.718} & \underline{2.076} \\
    Weighted (mean $N=5$) & \textbf{0.797$\pm$0.131} & \textbf{19.18$\pm$1.80} & \textbf{0.987$\pm$0.536} & \textbf{1.373} \\
    \bottomrule
  \end{tabularx}
\end{table}

Against compute-matched random augmentation, weighted mixture improved SSIM by 0.019 (0.013--0.025), PSNR by 0.95\,dB (0.66--1.27), and MSE by 0.0026 (0.0019--0.0034). All three pairwise contrasts within each metric remained significant after Holm correction (largest adjusted $p=0.0178$).

In the post-hoc all-$N=5$ analysis, fixed-mask performance was nearly unchanged. Weighted mixture beat fixed mask in 66/75 cases on each metric; the three weighted-versus-fixed contrasts had Holm-adjusted $p<2.5\times10^{-11}$, and all nine pairwise metric contrasts remained significant (largest adjusted $p=0.018$). Thus, unequal trajectory count does not explain the selection of weighted mixture over fixed mask.

\begin{figure}[!t]
  \centering
  \includegraphics[width=1\textwidth]{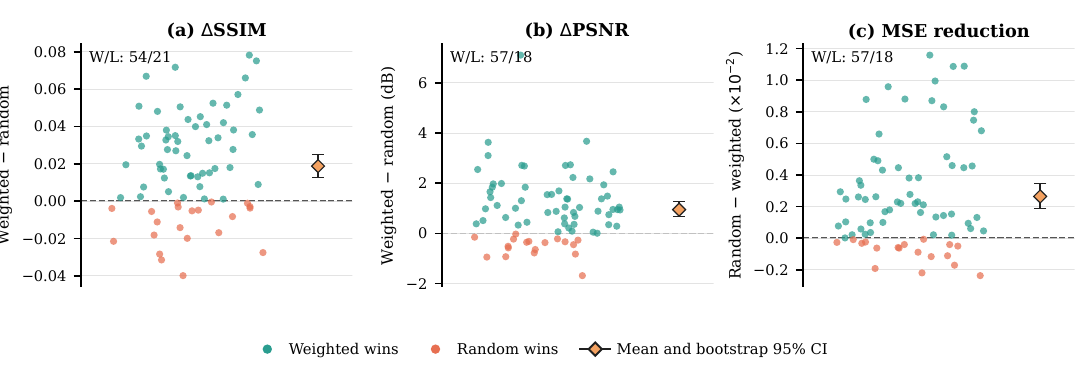}
  \caption{Compute-matched paired effects of weighted mixture versus random augmentation across the 75 internal model-selection cases. Positive values favor weighted mixture. Points represent cases; diamonds show mean paired effects with 100,000-resample case-bootstrap 95\% CIs. W/L denotes wins/losses.}
  \label{fig:confirmation-effects}
\end{figure}

Fig. \ref{fig:confirmation-effects} shows that the compute-matched gain over random augmentation was broadly distributed: weighted mixture won 54/75 cases for SSIM and 57/75 for PSNR and MSE. It improved all metrics in 51 cases, lost all three in 15, and had mixed effects in 9; across pipelines, it was best on all metrics in 46 cases.

A post-hoc exploratory analysis stratified weighted-minus-random differences by tertiles of the provided healthy scoring-mask volume ($n=25$ each), excluding the complete inpainting mask. For small, medium, and large scored holes, respectively, $\Delta$SSIM was $+0.012/ +0.023/ +0.022$, $\Delta$PSNR was $+1.08/ +0.88/ +0.88$\,dB, and $\Delta$MSE was $-0.0020/ -0.0031/ -0.0028$. SSIM and MSE gains were smaller for small holes; we do not interpret medium--large differences.

\begin{figure}[!t]
  \centering
  \includegraphics[width=1\textwidth]{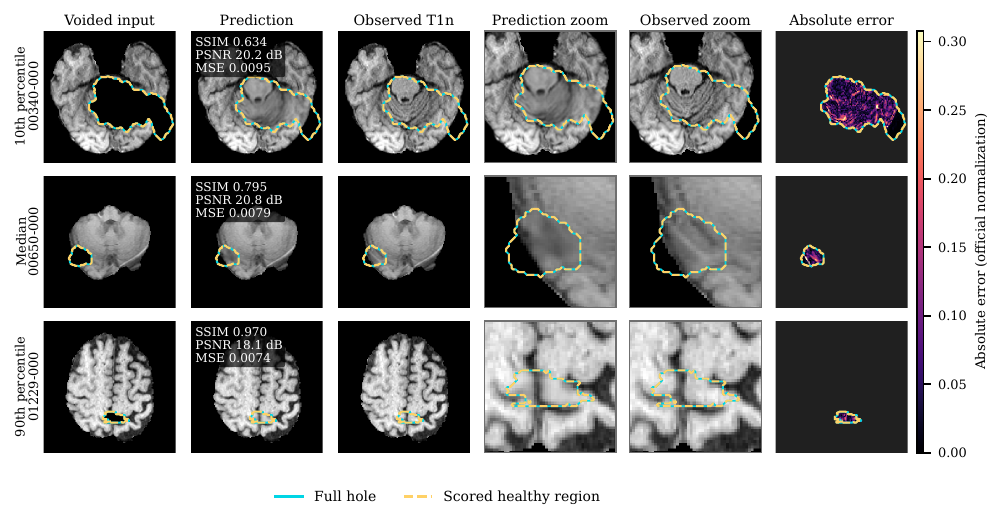}
  \caption{Weighted-mixture mean-$N=5$ reconstructions at the empirical 10th, 50th, and 90th SSIM percentiles. Cyan denotes the complete inpainting region and dashed yellow the scored healthy region. Error is shown only within the scored region on a shared clipped scale.}
  \label{fig:qualitative-reconstructions}
\end{figure}

Fig. \ref{fig:qualitative-reconstructions} illustrates case difficulty: the low-SSIM reconstruction preserves gross structure but has localized intensity and boundary mismatch, the median retains residual error, and the high-SSIM reconstruction closely matches the target. Because the observed T1n may remain pathological elsewhere in the complete inpainting region, these examples do not establish counterfactual accuracy.

The selected weighted-mixture mean-$N=5$ pipeline achieved SSIM $0.772\pm0.119$, PSNR $20.89\pm3.27$\,dB, and MSE $0.0098\pm0.0054$ on the 219-case official BraTS 2026 validation set. This organizer result is not a three-arm comparison.

\section{Discussion} \label{sec:discussion}

Weighted mixture is the strongest pipeline on the 75-case internal model-selection set and is selected for organizer evaluation. Random and weighted mixture differ only in healthy-mask policy (Sec. \ref{sec:exp:data}) and share inference, split, noise trajectories, and evaluator, so their comparison is compute-matched. Their training policies nevertheless differ jointly in mask families, volume sampling, transforms, current-case eligibility, epoch refresh, and fallback, so the evidence supports selecting the complete policy rather than its 80/15/5 proportions or any single component. Improvements in most cases argue against an outlier-only effect, but 15 all-metric losses show that the method is not uniformly superior. Because these 75 cases determine the selected arm, their estimates and hypothesis tests describe the internal selection comparison rather than independent post-selection confirmation.

The frozen fixed-mask pipeline uses $N=1$ because mean $N=3$ gave only a small development-rank gain at triple cost and mean $N=5$ was worse; its comparisons with the augmented pipelines therefore mix training-policy and inference-compute differences. The post-hoc analysis in Sec.~\ref{sec:results:confirmation} equalizes inference at mean $N=5$ and leaves the weighted-vs-fixed advantages intact, removing unequal inference compute as an explanation while retaining $N=1$ as the primary fixed-mask policy.

Mean-$N=5$ inference requires five 1000-step trajectories per case. On NVIDIA A100 40\,GB GPUs, steady-state wall-clock time was 5.8 minutes per case at $N=1$ and 29 minutes at mean $N=5$ ($4.9\times$), measured between consecutive saved predictions within each deterministic shard and excluding model startup. Faster wavelet inpainting \cite{durrer2026fastwdm3d} should therefore be evaluated under the same cohort and protocol.

Evaluation covers only artificially removed healthy voxels; since the observed T1n remains pathological in $m_u$, neither quantitative metrics nor qualitative references establish patient-specific counterfactual correctness in the real tumor region. Plausible reconstructions may still hallucinate incorrect anatomy, and clinical validity remains untested.

The case-keyed split is not fully patient independent: the optimization pool contains sibling timepoints for 3/25 development and 11/75 model-selection cases. Among the 64 model-selection cases patient-disjoint from optimization, weighted/random/fixed SSIM is 0.810/0.791/0.770, PSNR is 19.27/18.33/17.93\,dB, and MSE is 0.0092/0.0117/0.0128. The ordering persists, although optimization-pool siblings may influence development-stage selection.

\subsection{Limitations} \label{sec:discussion:limitations}

Limitations include one seed per arm; five-case, single-trajectory checkpoint selection, risking selection bias and noise-seed variance; transfer of terminally selected inference policies to arm-specific checkpoints; and no comparable external baseline on BraTS 2026, precluding claims over prior methods. Fixed-mask mean-(N=5) and void-size analyses were post hoc and do not replace the frozen protocol; only the weighted mixture underwent official 219-case evaluation, which cannot confirm the three-arm ordering. Future work should add seeds and baselines, test faster samplers and patient-level splits, and explore anatomy-aware plausibility.

\section{Conclusion} \label{sec:conclusion}

We presented CATCH, a conditional 3D Haar-wavelet diffusion framework for healthy brain-tissue inpainting, combining full-volume wavelet-domain diffusion, tumor-excluded supervision, hole-focused loss weighting, and hard compositing of the observed image. Within CATCH, weighted-mixture augmentation outperformed compute-matched tumor-component augmentation on the 75-case internal model-selection set, with the same ordering in the patient-disjoint analysis. The selected weighted-mixture mean-$N=5$ pipeline achieved SSIM 0.772, PSNR 20.89\,dB, and MSE 0.0098 on the official 219-case validation set.

\begin{credits}
\subsubsection{\ackname}
This project is supported by the Pioneer Centre for AI, funded by the Danish National Research Foundation (grant P1). Data used in this publication were obtained as part of the Challenge project through Synapse ID (syn74274097).

\subsubsection{\discintname}
The authors declare no competing interests.
\end{credits}

\bibliographystyle{splncs}
\bibliography{references}

\end{document}